\documentclass{article}

\usepackage[preprint,nonatbib]{neurips_2021}

\usepackage[utf8]{inputenc} 
\usepackage[T1]{fontenc}    
\usepackage{hyperref}       
\usepackage{url}            
\usepackage{booktabs}       
\usepackage{amsfonts}       
\usepackage{nicefrac}       
\usepackage{microtype}      
\usepackage{xcolor}        
\usepackage[english]{babel}
\usepackage[utf8]{inputenc}
\usepackage{listings}
\usepackage{color}
\usepackage{algorithm}
\usepackage{algorithmic}
\usepackage{todonotes}
\usepackage{siunitx}
\usepackage{authblk}
\usepackage{subcaption}
\usepackage{bbm}
\usepackage{amsmath}

\title{Using mixup as regularization and tuning hyper-parameters for ResNets}

\author{%
    Venkata Bhanu Teja Pallakonda \\
    
    Texas A\&M University \\
    College Station, TX, 77801 \\
    \texttt{bhanu@tamu.edu} \\
}

\begin{document}

\maketitle

\begin{abstract}
 While novel computer vision architectures are gaining traction, the impact of model 
 architectures is often related to changes or exploring in training methods. Identity mapping based architectures ResNets \cite{resnets} and DenseNets \cite{densenets} have promised path breaking results in the image classification task and are go to methods for even now if the data given is fairly limited. Considering the ease of training with limited resources this work revisits the ResNets and improves the ResNet50 \cite{resnets} by using mixup data-augmentation as regularization and tuning the hyper-parameters.
\end{abstract}

\section{Introduction}

The performance of a vision model is dependent on both model architecture and training methods. With the findings of attention \cite{attention} based architectures the vision transformers \cite{vit} achieved the state of art results on the image classification tasks. But training time is expensive and we need to large data-sets for learning. Vision transformers are trained using JFT 300M \cite{jft300m} which happens to be private data. Keeping in view of all these shortcomings and if the provided dataset is not large enough we go back to skip connection and Identity mapping based ResNet \cite{resnets} and DenseNet \cite{densenets} architectures which performs reasonably well till date on Image classification tasks.

We use the idea of data augmentation task, mixup as a regularization \cite{mixup} task to improve our test set validation error (there by increasing accuracy) which would also help us to do better classification if we want to classify a new image that completely out of domain of our train set. We also do Bayes hyper-parameters sweep using wandb \cite{wandb} to find the best hyper-parameters for our model after applying this regularization task. 

\section{Methodology}
Since the introduction of AlexNet \cite{alexnet} on ImageNet \cite{Imagenet}
 various methods have  been proposed to further improve image recognition performance. These improvements typically occur on architecture and training methods.

 \subsection{Architecture}

\begin{figure}[]
  \centering
  \includegraphics[scale=.5,angle=90,origin=c]{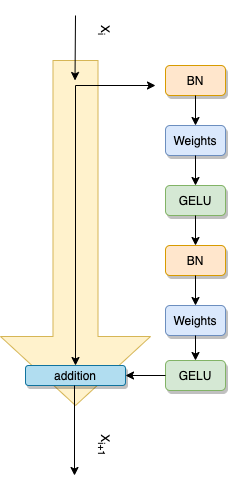}
  \caption{Skip connection block with tweaked activations and order of the activations.}
  \label{fig:skipconnection}
\end{figure}

 We experimented and used ResNet50 \cite{resnets} architecture with the preactivations \cite{preactivations} at every bottleneck block \cite{resnets} and gelu \cite{gelu} non linear activation functions. The skip connection block used in the ResNet is shown in Figure \ref{fig:skipconnection}. We use this skip connection as basic block for the ResNet50 architecture.

\subsection{Augmentations}
Data augmentation is an important techinque that will be very handy to imporove out architecture and for improving prediction on out of domain data. In this method we follow mixup (Deatiled explnation in section \ref{mixup}), cropping and random horizontal flipping. As the images dimmensions in CIFAR-10 \cite{cifar10} is small the cropping and flipping would be very useful.

\section{Exploring mixup and using it as regularization factor}
\label{mixup}
This section describes how the mixup is performed, how to calculate the loss after mixup for the backpropagation tasks, and how could it can be related it to regularization.

\subsection{Mixup calculation}

\begin{figure}[]
  \centering
  \includegraphics[scale=.2]{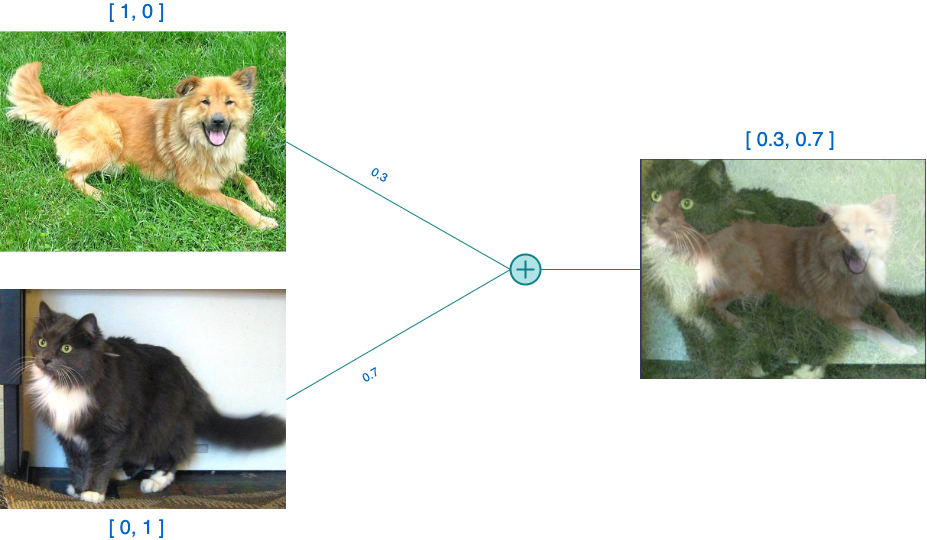}
  \caption{Mixing 2 images ($\lambda=0.3$)}
  \label{fig:mixup}
\end{figure}

As show in Figure \ref{fig:mixup} we overlay one image from one class to a different image in another class with an an randomly generated overlay factor of $\lambda$ from beta distribution \cite{betadist} during training. A simple algorithmic approach is found below in algorithm \ref{algo:mixup}.

\begin{algorithm}
	\caption{mixup mechanism}
	\label{algo:mixup}
	\begin{algorithmic} 
	    \STATE $ s = \  inputs \ size $
		\STATE $\lambda = \ np.random.beta(alpha, alpha, s) $
		\STATE $index = \ np.random.permutation(s) $
		\STATE $x_1,x_2 = \ inputs, inputs[index, :, :, :]$
		\STATE $ y_1, y_2 = \ onehot(targets, numclasses), onehot(targets[index,], numclasses) $
		\STATE $ final\_input = \ \lambda*x_1 + (1-\lambda)*x_2  $
		\STATE $ final\_target = \ \lambda*y_1 + (1-\lambda)*y_2 $
	\end{algorithmic}
	
\end{algorithm}


\subsection{Cross entropy loss calculation for mixup}
\begin{figure}[]
  \centering
  \includegraphics[scale=.4]{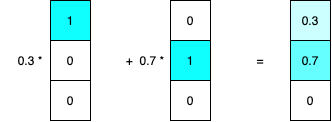}
  \caption{Truelable for mixup}
  \label{fig:truelabel}
\end{figure}

As shown in Figure \ref{fig:truelabel} we take the new true label $\boldsymbol y$ and compute the cross entropy loss aganist the predicted probabilities $\hat{\boldsymbol y}$ of our model. We follow Algorithm \ref{algo:mixup_loss} for calculating the effective cross entropy loss for the mix up input images to the model. 
\begin{equation}
  CE({\boldsymbol y}, \hat{\boldsymbol y}) = - \sum^{\textrm{N}_{c}}_{i=1} y_{i}\log(\hat{y}_{i})
  \end{equation}

\begin{algorithm}
	\caption{mixup loss calculation}
	\label{algo:mixup_loss}
	\begin{algorithmic} 
		\REQUIRE $\lambda \ from \ mixup \  algorithm $
		\REQUIRE $final\_input \ from \ mixup \  algorithm $
		\REQUIRE $final\_target \ from \ mixup \  algorithm $
		\STATE $ x_p = model(final\_input) $
		\STATE $ x_p = torch.log(torch.nn.functional.softmax(x_p, dim=1).clamp(1e-5, 1)) $
		\STATE $ final \ loss = \ - torch.sum(x_p * final\_target)$
	\end{algorithmic}
	
\end{algorithm}

\subsection{Relating to regularization}
\begin{figure}[]
  \begin{subfigure}{0.49\textwidth}
    \includegraphics[width=\linewidth]{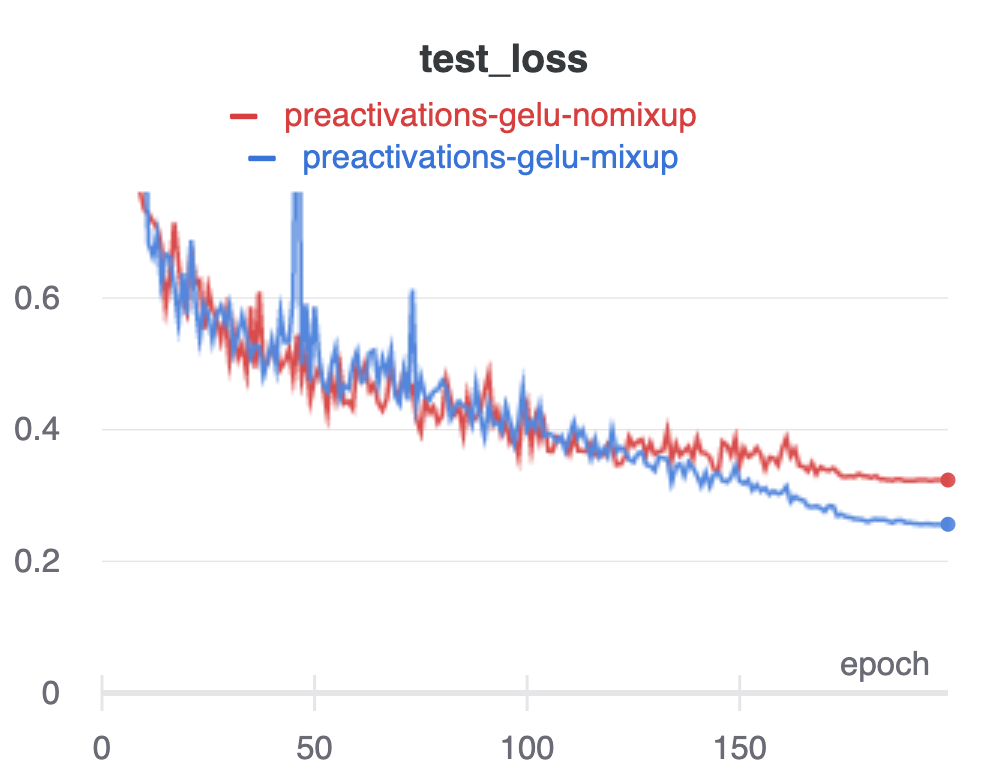}
    \caption{Test loss} \label{fig:rega}
  \end{subfigure}%
  \hspace*{\fill}   
  \begin{subfigure}{0.49\textwidth}
    \includegraphics[width=\linewidth]{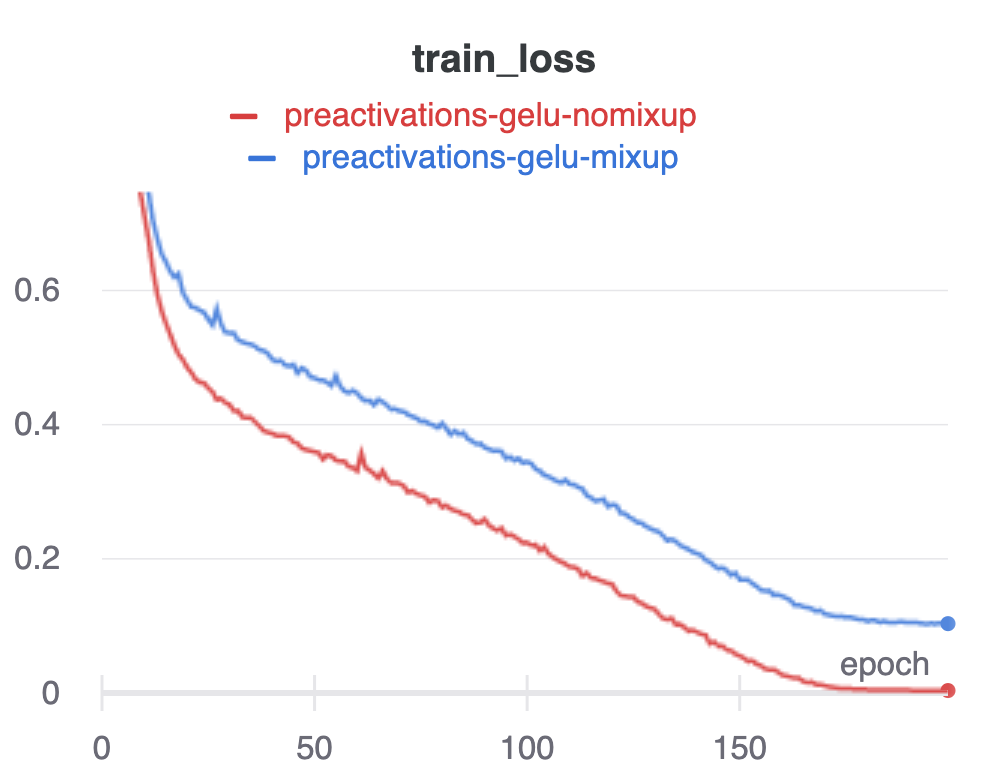}
    \caption{Train loss} \label{fig:regb}
  \end{subfigure}%
\caption{Losses with and without mixup} \label{fig:reg}
\end{figure}

For experimentation we use ResNet50 \cite{resnets} with GELU \cite{gelu} non-linear activation function and pre-activation's \cite{preactivations} at every bottleneck layers. With the same given set of hyperparameters and traing the two different models as shown in Figure \ref{fig:rega} we could clearly see that loss on the test set is decrease by almost a factor of 1.5 (Also increasing the test accuracy) which could be  helpful for predicting the images that are not the same domain of our train set.
This could be related to regularization. We can see the the train loss in Figure \ref{fig:regb} is higher for this method as we are introducing more uncentrainity to the training samples by mixing up and and forcing our model to do image classification by trying to maximize the single class probability. 

\section{Hyperparameter tuning}
\begin{figure}[]
  \centering
  \includegraphics[width=\linewidth]{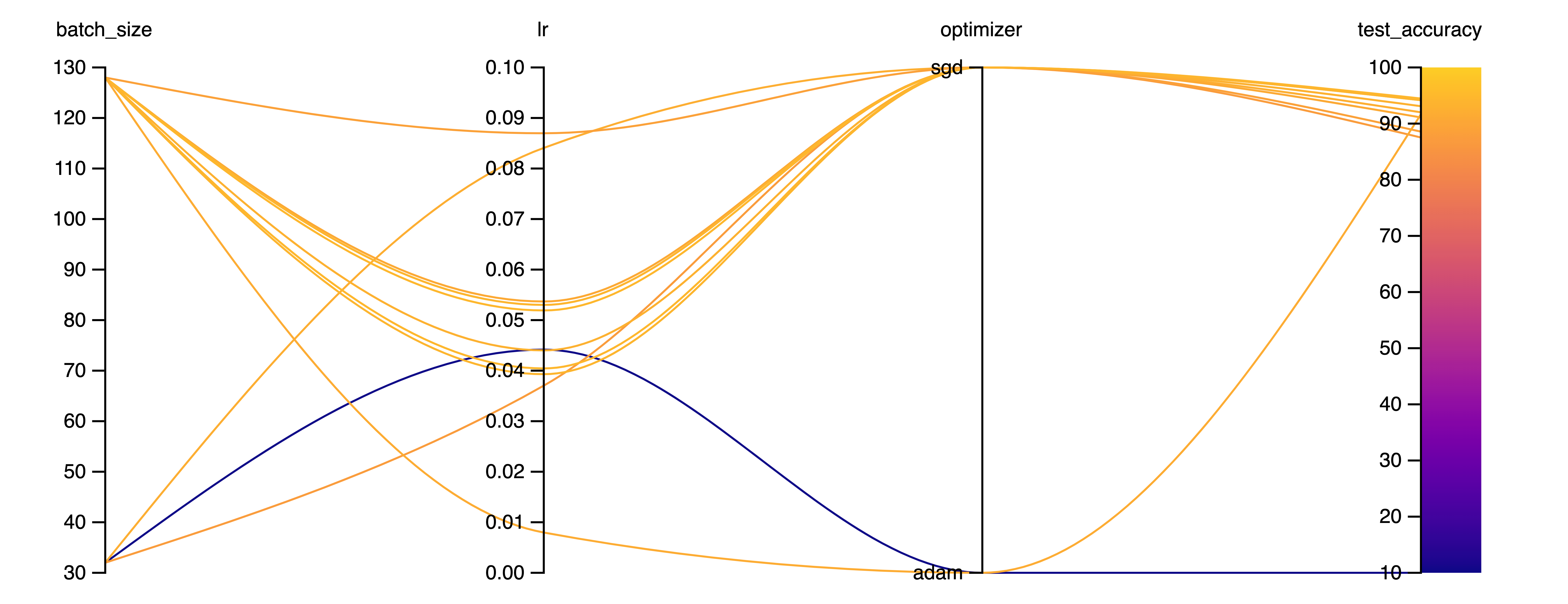}
  \caption{Test accuracies over a range of hyperparameters}
  \label{fig:sweepwandb}
\end{figure}

\begin{figure}[]
    \centering
    \includegraphics[width=\linewidth]{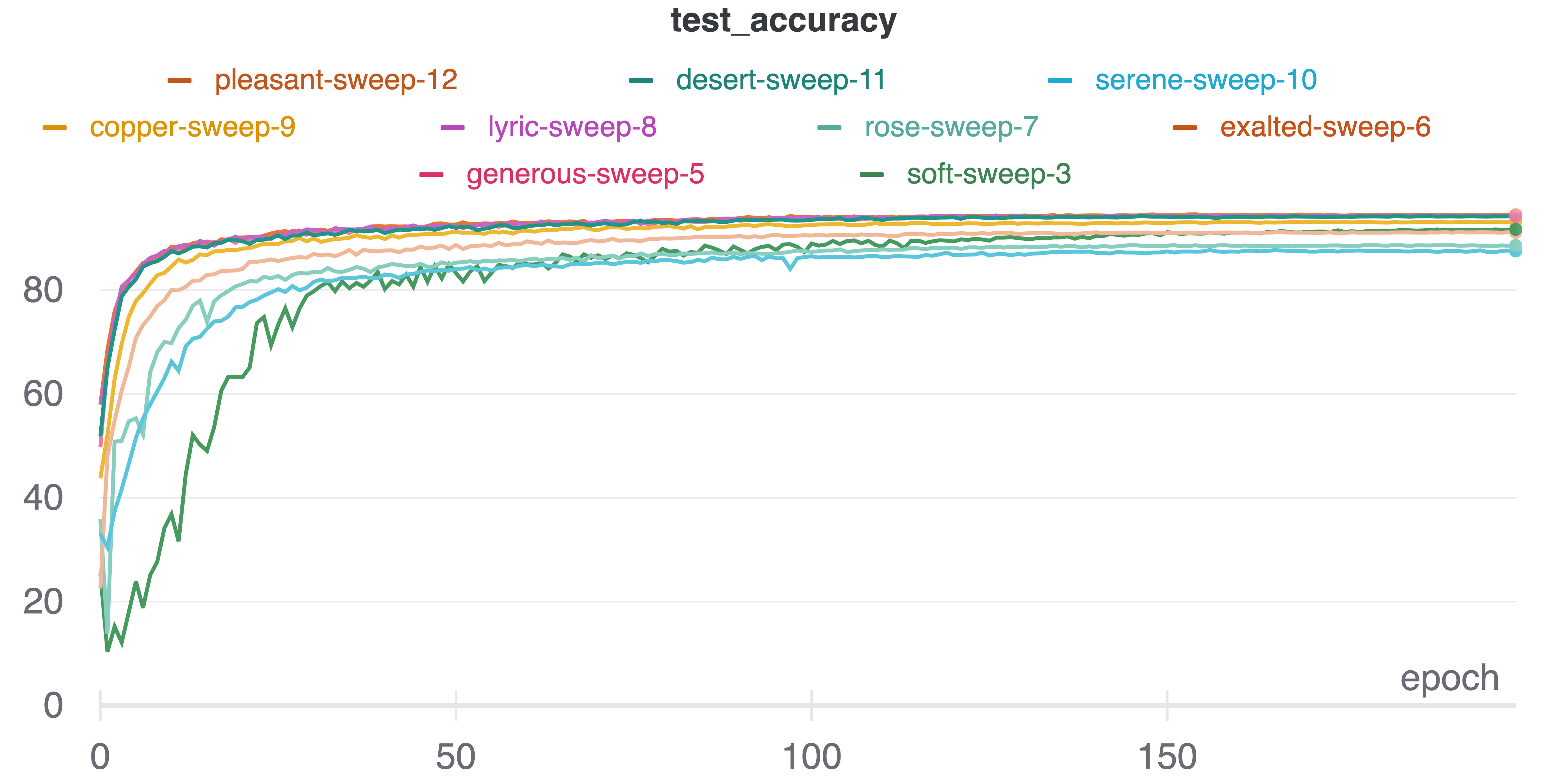}
    \caption{Speed of convergence} \label{fig:sweepa}
\end{figure}

\begin{figure}[H]
  \centering
  \includegraphics[width=\linewidth]{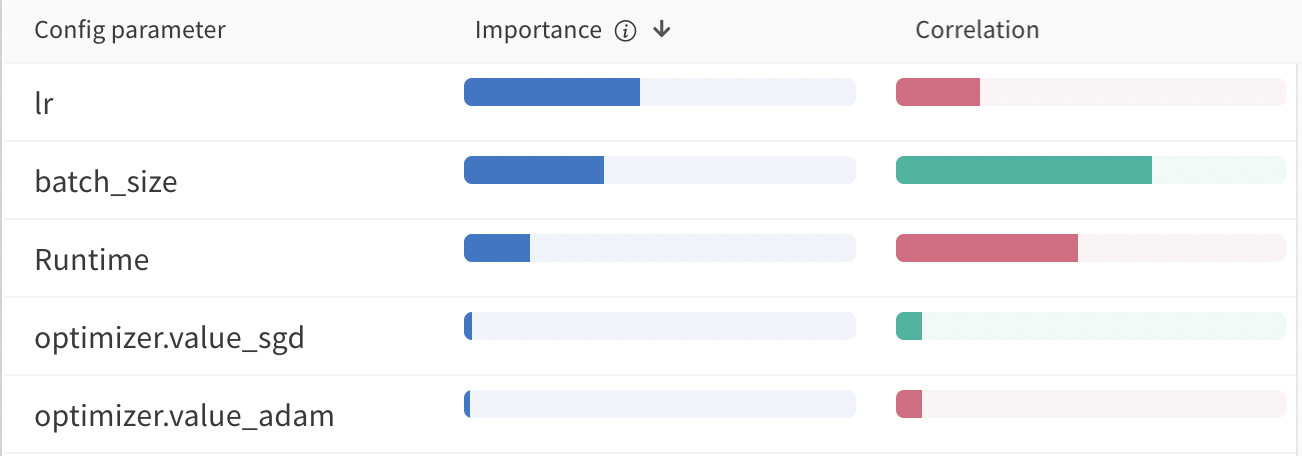}
  \caption{Hyperparameters correlation with test accuracy} \label{fig:sweep_correlation}
\end{figure}

Using weights and biases \cite{wandb} parameter sweeps are done with hyperband \cite{li2018hyperband} early stopping. The below figures \ref{fig:sweepwandb} illustrate the test accuracies over a range of various hyper-parameters and Figure \ref{fig:sweepa} was plotted to check the speed on convergence of the network. From figure \ref{fig:sweep_correlation} we could see the correlation of hyperparameters with test accuracies and their importantance in the sweep.

After runnng the experimentes the optimal hyperparameters are shown in Table \ref{hyperparameter}

\begin{table}[]
  \caption{Optimal hyperparameters after experimentation}
  \label{hyperparameter}
  \centering
  \begin{tabular}{@{}|l|l|@{}}
    \toprule
    Hyperparameter              & Value                          \\ \midrule
    Learning rate           & 0.05                                \\ \midrule
    Batch size          & 128                                \\ \midrule
    Optimizer         & SGD                               \\ \midrule
    Learning rate scheduler         & CosineAnnealing $T_m = 200$ and epochs = 200                                \\ \bottomrule
    \end{tabular}
    \end{table}

\section{Results}
This method achieves an overall accuracy of \textbf{94.57\%} on the CIFAR-10 dataset \cite{cifar10}. Below Table \ref{comp-table} shows comparisions with the exisitng ResNet architectures. The ResNet50 could perform well when compared to other higher dept architectures by introducing this idea of mixup.

\begin{table}[H]
  \caption{Classification error (\%) on the CIFAR-10 test set}
  \label{comp-table}
  \centering
  \begin{tabular}{@{}|l|l|@{}}
    \toprule
    network              & error(\%)                            \\ \midrule
    resnet-50            & 6.97                                 \\ \midrule
    resnet-110           & 6.61                                 \\ \midrule
    resnet-164           & 5.93                                 \\ \midrule
    resnet-1001          & 7.61                                 \\ \midrule
    \textbf{This method} & {\color[HTML]{000000} \textbf{5.43}} \\ \bottomrule
    \end{tabular}
    \end{table}
\bibliographystyle{IEEEtran}
\bibliography{main.bbl}        
\end{document}